# Design Rule Checking with a CNN Based Feature Extractor

*Luis Francisco, Tanmay Lagare, Arpit Jain, Somal Chaudhary, Madhura Kulkarni,
Divya Sardana, **W. Rhett Davis and ***Paul Franzon
[*lsfranci,**wdavis,***paulf]@ncsu.edu
North Carolina State University
Raleigh, NC, USA

## ABSTRACT

Design rule checking (DRC) is getting increasingly complex in advanced nodes technologies. It would be highly desirable to have a fast interactive DRC engine that could be used during layout. In this work, we establish the proof of feasibility for such an engine. The proposed model consists of a convolutional neural network (CNN) trained to detect DRC violations. The model was trained with artificial data that was derived from a set of 50 SRAM designs. The focus in this demonstration was metal 1 rules. Using this solution, we can detect multiple DRC violations $32x$ faster than Boolean checkers with an accuracy of up to 92%. The proposed solution can be easily expanded to a complete rule set.

## 1 INTRODUCTION

Design Rule Checking (DRC) is the process of checking that design geometry satisfies a set of layout rules. The ultimate driver for doing a DRC check is to ensure that, when fabricated, the design will achieve a high yield, as limited by the manufacturing tools and steps. They are generally specified as a set of logical rules. For example, a wire has to be greater than a certain minimum width, among others. Typically, the DRC is referred to as a "final sign-off" check. The entire chip is run through the design rule checker, taking a long time to do so.

The design rule checking process is becoming extremely complex in advance technology nodes, especially below $28nm$. Every foundry and each technology node requires unique DRC checks. This difficulty is related to the complexity of the lithography process and the minuscule depth-of-focus tolerance. Another reason for this is the inclusion of complex rules introduced by the need to support double patterning [5]. According to EDA vendors, the number of rules has grown from a few hundred in $60nm$ nodes to thousands of rules in $7nm$ nodes. A data-driven approach, such as machine learning (ML) for DRC, can help with its complexity because it can capture effects that go beyond physical parameters on the designs.

Since there has been an increase in intricacies in rules, the speed of the rule checker has become a critical point in layout generation of custom circuits. Before advanced nodes, layout engineers could recognize patterns and fix DRC violations with relative ease. Nowadays, the design engineer waits a long time for the final sign-off DRC checker to run. The engineer then fixes the problems flagged, often only to introduce new ones. This iterative process is becoming unacceptable due to the slow checkers and rules complexity.

Machine learning can help to speed up the design rule checking process because a neural network does not have to iterate over thousands of Boolean operations.

Some work in improving the DRC rules definition, and the checking process can be found in the research state of the art. A common theme in recent research on DRC is to move from model-based approaches to data-driven solutions. An existing methodology to redefine DRC rules is DRC+ [3]. DRC+ defines the rules based on the identification of hotspots using pattern printability simulation when a problematic pattern is identified; a DRC+ rule is crafted to mitigate its effect. DRC+ predates the machine learning revolution. It relies more on pattern matching. Another data-driven approach is found in OpenDFM [2]. In contrast with conventional DRC, OpenDFM is not a binary checker that provides only a pass/fail output; it offers more information about a set of parameterized patterns.

Currently, ML is a mature enough topic to be implemented to solve electronic design automation problems. Among those problems, hotspot detection shares some similarities with DRC. Some works apply ML to lithography [11, 15, 16] and chemical mechanical polishing (CMP) hotspot detection [4]. In [16], a support vector machine is trained with geometrical and other critical features. In contrast, [11, 15] proposed deep learning without any previous feature extraction; the model itself performs the feature extraction. In our proposed work, we use a similar deep learning approach; however, it goes beyond hotspot detection and focuses on DRC violation detection.

For the purposes of detecting design rule violations, some machine learning approaches have been tried [7, 12, 14, 17]. In [12], a machine learning method is used to detect detailed routing short violations. A neural network is trained to predict short violations using features extracted from a placed and routed netlist. In [14] ML is used to predict routability. This work uses a convolutional neural network (CNN) to predict ASIC routablity and the potential for DRC violations, without identifying the violations. The model was trained with features from each stage of the physical design flow (floorplanning, placement, and routing). Another model to predict routability and to detect the potential presence of DRC violations is presented in [7]; this approach uses random forest models. The work in [17] focuses more on a neural network (to predict the presence of violations) and compares the results with a random forest predictor. In [8], ML is used to predict the number of DRC violations after each routing stage and similar to previously referenced works, without locating and identifying them. These solutions focus on predicting DRC violations in global and detailed routing; this has relevance only in a physical design flow for digital ICs. [9] introduced a hotspot predictor using CNN architecture (J-Net) for DRC hotspots with mixed resolution features. This work results in a binary classifier to identify layouts containing hotspots and clean layouts.

In contrast with the works previously described, we are proposing a deep learning method based on a convolutional neural network to detect multiple DRC violations. This solution is able to detect and identify the violations, not only label layouts areas. The model uses raw images from multiple layouts as training data. The images contain labels for DRC violations randomly inserted in the layout. By using the layout images as input, the model is trained without any dependency on physical parameters. The feature extraction is performed by the model using the convolutional layers. This method is expandable to any number of rules, including the most complex ones and extrapolated to different technology nodes. It can be used not only for digital IC design flows but for custom design layouts. An approach like this can help close the gap between DRC and design for manufacturing (DFM) by proving the ability to train the model with Fab data and with feedback from designers.

The rest of this document is organized as follows: section 2 describes the proposed deep learning model, section 3 presents how the training dataset was generated and 4 introduces the metrics to evaluate the model. Lastly, section 5 shows the experimental results followed by final remarks and future work in 6.

## 2 PROPOSED DEEP LEARNING MODEL

One advantage of a deep learning model is that it does not require a previous feature extraction method. This characteristic removes the dependency on the technology node and the human iteration experience. This advantage increases the number of features that can be extracted by the model compared to a feature extractor based on designer experience.

A deep learning model consists of various learning layers or hidden layers. The learning layers extract features in the form of data representations of the input. Typically in between each learning layer, a down-sampling layer is used to reduce the size of features extracted. Followed by the hidden layers, a fully connected layer discriminates the learned features to map them to a classification output probability. After defining the model structure, the training process uses labeled data to generate predictions and minimize the prediction error. Figure 1 illustrates how the training process is done.

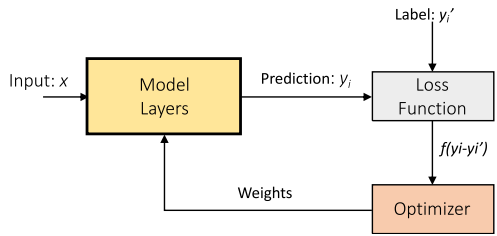

Figure 1: Deep learning model training flow.

The training process can be seen as an optimization problem [4]; thereby, we must define the loss function and the optimizer. Those parameters have an impact on model performance.

## 2.1 Convolutional Neural Network

The convolutional layers perform the feature extraction. Those features are extracted by doing a spatial convolution operation between an input $x$ and a kernel $h$. This operation is described as follows,

$$x[n] * h[m] = \sum_{k_1}^{N} \sum_{k_2}^{M} x[k_1, k_2] \cdot h[n - k_1, m - k_2], \quad (1)$$

where $N$ and $M$ are the dimensions of the input and the kernel, respectively. The initial values for the kernel are randomized and tuned in on every training iteration. The kernel values represent the weights in Figure 1. The size and number of kernels for each layer will determine the number of trainable parameters for the convolutional layer.

In a CNN, each convolution operation is followed by an activation function. For the model presented in this work, we use a rectifier linear unit (ReLU) function. The ReLU function takes only the positive part of its input. We explored other activation functions, but ReLU performed the best. This function is easily implemented. The same activation function is used for the output intermediate fully connected layer providing a classification probability.

## 2.2 Loss Function and Optimizer

As we are working in a classification problem and the model prediction is a probability, the loss function selected is cross-entropy [13]. This loss function provides values proportional to the difference in the prediction probability and the target labels.

We explored different gradient descent based optimizers having the best result with *RMSprop* introduced in [6]. *RMSprop* has been used in other deep learning models for electronic design automation applications [4]. This optimization algorithm reduces the learning for a given weight by the root mean square (RMS) of the prior gradients. Each weight $w(t)$ is given by,

$$w(t) = w(t-1) - \alpha \frac{\eta}{\sqrt{S_q(w,t)}} \partial E, \quad (2)$$

with $S_q(w, t)$ given by

$$S_q(w, t) = \alpha S_q(w, t-1) + (1 - \alpha) \left( \frac{\partial E}{\partial w(t)} \right)^2, \quad (3)$$

where $E$ is the loss function, $\eta$ is the learning rate and $\alpha$ is a parameter.

## 2.3 Model Structure Selection

By putting together all the components previously described, we built a deep learning model with a convolutional neural network to perform the feature extraction. The model takes as input the layout clip images and provides a classification output probability. By taking the maximum probability, the layout clips are labeled as a DRC violation, containing several violations, or DRC clean.

Figure 2 summarizes the model structure and illustrates what is expected in each stage. The model settings are used to evaluate different model structures are as follows:

- The **input** image **size** is set according to pixels and metal pitch. For example, 1 pixel corresponds to $1nm^2$.

- The number of **hidden layers**/**Convolutional layers** is set to avoid overfitting and underfitting. In the same way, the number and size of the filters.
- The **resampling layers** are added to reduce the output size of the convolutional layers. We use max pooling for the resampling layers.
- The **intermediate layer** or **fully connected** layer maps the extracted features to an output probability. The size $M$ of this layer has an impact on the total number of trainable parameters for the model.
- The **output layer** contains a dimension $d$ that is given for the combination of all the possible DRC violations analyzed. The label is assigned, taking the maximum probability for a class.

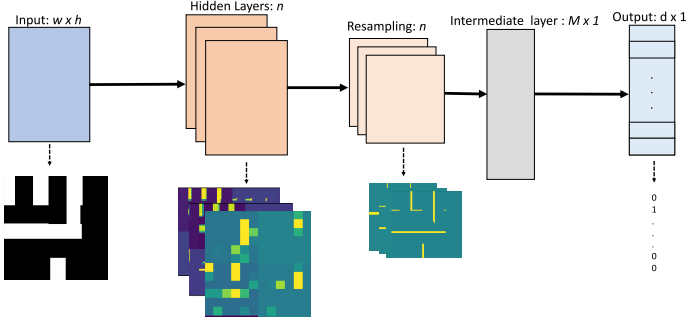

Figure 2: Propose model structure.

Table 1 shows the settings to construct two different model structures. The first model is to detect 1 DRC violation. This model contains 159,746 trainable parameters with four feature extraction layers. The second model is to target 3 DRC violations. The number of trainable parameters is increased to 1,673,256 as it should detect up to 8 classes (each violation and the possible presence of many rules). These parameters determine the model complexity, which impacts the model performance. To select the parameters we need to monitor overfitting and underfitting. The number of parameters has more impact on the training time than inference time.

Following the previous model structure, the model complexity can be set to target any number of rules. A more complex model will not always result in better performance. A deep learning model is as good as the diversity or generalization in the training dataset.

## 3 DATASET

Data that includes a large number of DRC free layouts along with layouts containing errors is necessary to create the model. A dataset with over 250,000 layout clips, is created to train, validate, and test the model. The dataset consists of images basically from SRAM layouts using $15nm$ the public design kit [1]. The initial designs are DRC free, thus creating the need to insert random DRC violations. The training samples for all the test performed to the model is about 80%, the validation 15% and the testing the remaining 5%.

### 3.1 Dataset Generation Process

The process of generating the dataset is in Figure 3. After selecting a layout with no DRC violations, the following steps are performed:

Table 1: Model parameters example for 1 and 3 DRC violations.

| Model Setting | Target Rules | |
|---|---|---|
| | 1 DRC | 3 DRC |
| Input | Size: 200x200, 1px/1nm | Size: 200x200, 1px/1nm |
| Conv2D #1 | 32: 3x3 filters, params: 320 | 16: 3x3 filters, params: 160 |
| MaxPool2D #1 | Downsample: 2x2, resize by 2 | Downsample: 2x2, resize by 2 |
| Conv2D #2 | 16: 3x3 filters, params: 4,624 | 32: 3x3 filters, params: 4,640 |
| MaxPool2D #2 | Downsample: 2x2, resize by 2 | Downsample: 2x2, resize by 2 |
| Conv2D #3 | 16: 3x3 filters, params: 2,320 | 32: 3x3 filters, params: 9,248 |
| MaxPool2D #3 | Downsample: 2x2, resize by 2 | Downsample: 2x2, resize by 2 |
| Conv2D #4 | 32: 3x3 filters, params: 4,640 | 64: 3x3 filters, params: 18,496 |
| MaxPool2D #4 | Downsample: 2x2, resize by 2 | Downsample: 2x2, resize by 2 |
| FC | Size: 128x1, params: 147,584 | Size: 128x1, params: 1,638,656 |
| Output | Dimension: 2, pasrams 258 | Dimension: 8, params: 2,056 |
| Total Params | 159,746 | 1,673,256 |

(1) Extract the metal layers where the DRC will be performed.
(2) Insert random variations in the layouts. Those variations are selected to create DRC violations. In total 5,000 variants of these designs were auto-generated using random parameters.
(3) Run layouts through a conventional DRC checker to verify that the variations create real rules violations. A mask is created with this information to label each clip as containing rule violations or not. Figure 4 illustrates a mask image.
(4) Crop the layout using a given windows size; in the result presented, the size of the window is $200x200nm$. The cropping window is moved with a stride of 150nm to avoid creating false rules.
(5) Convert each layout clip into an image. The relation to image pixels and layout dimensions is $1px$ to $1nm^2$.

To have an evenly balanced dataset, one sample with no DRC violations is added for each added clip containing violations. The clips in the dataset can include one or multiple DRC violations. By doing so, the model can identify each rule or diverse rule violations.

The seed to generate the dataset consists of fifty different SRAM designs generated by fifty teams in a graduate VLSI class. These were designed using the NCSU $15nm$ PDK [1], all of which were DRC free. In total, 5,000 variants of these designs were auto-generated using random parameters.

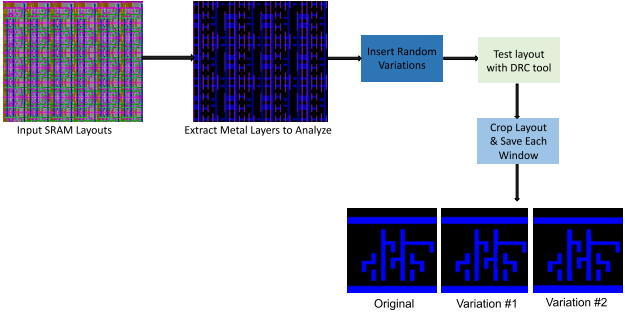

Figure 3: Dataset generation

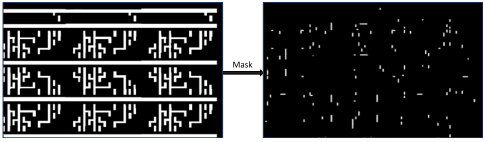

Figure 4: An example of a layout of 9 SRAM cells

## 3.2 Design Rules

The rules targeted in this work are the width and spacing standard design rules in the FreePDK15 [1]. In the dataset we included the following DRC violations:

- $M1.1$ - Minimum width of M1 metal layer is $28nm$.
- $M1.6$ - Minimum spacing of M1 to M1 be should be $36nm$.
- $M1.7$ - End-of-Line spacing of M1 to M1 should be $45nm$.

Table 2 shows how the dataset samples are labeled for three rules. The dataset includes samples with each independent rule, samples with combinations of them, and clean samples. The number of samples in each category remain the same.

Table 2: Labels in the dataset for 3 rules.

| Label | Description |
|---|---|
| DRC1 | Minimum width of M1 metal layer is $28nm$ |
| DRC2 | Minimum spacing of M1 to M1 should be $36nm$ |
| DRC3 | End-of-Line spacing of M1 to M1 should be $45nm$ |
| DRC12 | Sample containing DRC1 and DRC2 |
| DRC13 | Sample containing DRC1 and DRC3 |
| DRC23 | Sample containing DRC2 and DRC3 |
| DRC123 | Sample containing DRC1, DRC2 and DRC3 |
| NDRC | No DRC violation |

## 4 MODEL PERFORMANCE METRICS

After training the model, it is crucial to measure how well the model can perform; this is its ability to detect rules and how fast it can detect them. The metrics used most frequently are extracted from the classification confusion matrix. The confusion matrix contains true positives ($T_P$), false negatives ($F_N$), true negatives ($T_N$), and false positives ($F_P$).

The confusion matrix information can measure model performance in terms of recall and precision. The recall measures how well the model can detect DRC violations with respect to the $F_N$, while the precision compares the $T_P$ with $F_P$. The recall can be calculated as,

$$recall = \frac{T_P}{T_P + F_N}, \quad (4)$$

and the precision by,

$$precision = \frac{T_P}{T_P + F_P}. \quad (5)$$

As we know, it is more critical and costly from the design standpoint to miss DRC violations than to have false alarms; this is why we focus on getting a high recall when detecting them. This is similar to getting low false negative rate. The false negative rate is defined by,

$$FNR = \frac{F_N}{F_N + T_P}. \quad (6)$$

On the other hand, false positive rate is defined by,

$$FPR = \frac{F_P}{F_P + T_N}. \quad (7)$$

### 4.1 Testing New Layouts Process

We can also focus on another metric that determines how fast a model can perform in testing data and new layouts; this is also called the inference time. The process to test a new layout is shown in Figure 5 and consists of the following steps:

(1) Extract the metal layers to analyze.
(2) Crop the layout using a $200nm$ square windows, same as for the dataset using a stride of $150nm$.
(3) Convert layout clips to images and keep track of the coordinates in the layout.
(4) Test individual images and assign a class: DRC violation, multiple violations or, DRC free.
(5) Create a mask image and or layout mask to locate the violations on the layout.

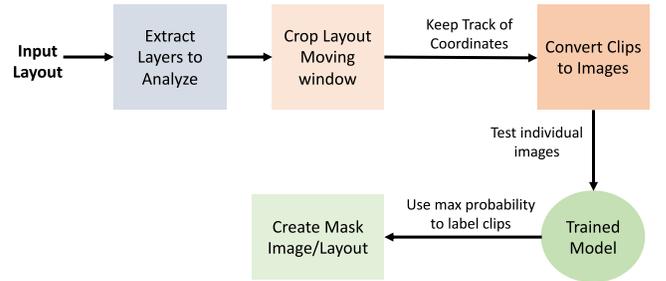

Figure 5: Inference flow to test new layouts.

Using the previously described flow, we can have an estimation of how fast our approach can perform compared to traditional DRC checkers. This approach can be used for any number of rules added to the model. This process allows the model to be included in an interactive DRC flow.

## 5 RESULTS

In this section, we present results that validate that it is possible to create a deep learning model using a CNN feature extraction to detect DRC violations. We include results that validate the model feasibility to detect one DRC violation and multiple violations.

A primary model was created to detect one DRC violation. In this initial experiment, the created model was trained with 50,000 layout clips containing many variations of the same DRC rule and 50,000 DRC clean. The model was validated with 15,000 samples and tested with 5,000 clips never seen by the model. Figure 6 shows the train and validation losses for the model with no overfitting or underfitting. The model requires around 7 minutes to be trained; note that the training is done in a 6 *CPUs* computer without using *GPU* acceleration.

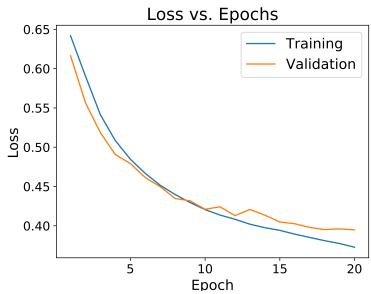

Figure 6: Training and validation loss for 1 DRC model.

Figure 7 shows the confusion matrix for the testing samples. The accuracy or recall to detect the DRC violation is 92%. The false positive rate is around 28%, and the false negative rate is around 8%. This result is encouraging as users deal better with false negatives than false positives.

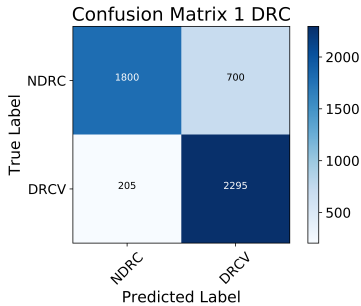

Figure 7: Confusion matrix for 1 DRC model.

Figure 8 presents the classification results for 4,800 samples containing 3 DRC violations. The confusion matrix shows that the model is classifying the majority of the samples correctly and that the false negatives are not as frequent as false positives. The samples used for the confusion matrix are samples never seen by the model.

For 3 rules, the diversity in the training dataset is not as high as for 1 rule. When more training iterations are added, the model resulted in overfitting and reduced the validation accuracy. Figure 7 illustrates how the validation loss of the model does not decrease while the training loss decreases. This behavior is mainly because, in the results presented, we use only SRAM designs as seeds to generate the dataset. We tried to reduce the model complexity by adding dropout and regularizer layers, but no significant improvement. To solve this issue, we proposed to increase the dataset diversity by including multiple types of designs and adding more rules variations. The community has studied this issue; its nature makes it challenging to achieve high detection rates to "never seen" data [10]. The work presented in [10] suggests novel augmentation methods to improve the classification rates.

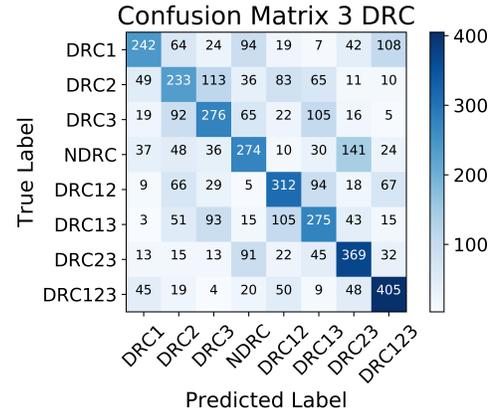

Figure 8: Confusion matrix for 3 DRC model.

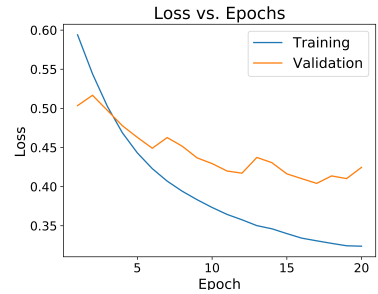

Figure 9: Training and validation loss for 3 DRC model.

Finally, the time to test a full layout $1000\mu m$ x $1000\mu m$ equivalent to 5,000 samples takes less than 13 seconds. We compared the speed of the CNN engine versus running the same rules set in a conventional DRC engine on one of the layouts. The CNN took 13 seconds versus 420 seconds for a conventional DRC engine $32x$ faster. Note that the conventional check was done with a runset containing only the violations implemented in the CNN model. While the accuracy of the proposed solution can be improved, the speed factor enables it to be used in the early stages of the design flow. Detecting and fixing violations early in the design reduces the design cycle time. In an industry application of this approach, the training data will come for multiple layouts and DRC patterns database, adding more diversity to the training data and including more complex rules.

## 6 CONCLUSIONS

In this work, we propose a new design rule checking approach using machine learning. This solution consists of a deep learning model with CNN as a feature extractor. The model is trained with layout images that can be labeled artificially or by designers. Using this model, we can detect DRC violations with an accuracy of up to 92% and up to $32x$ times faster than traditional boolean checkers. The speedup can help to increase designer productivity and reduce time to market by providing violations detection early in the design flow. It also has the potential for better automation in full custom layout. This approach is easy to expand to new technology nodes since the model itself extracts the features.

The proposed solution has the potential to be included in an interactive DRC engine based on machine learning models to help the layout engineer quickly converge on a DRC free layout. In future research, we will be focusing on adding more complex design rules and scale the work to a full DRC set.

## ACKNOWLEDGMENTS

The authors would like to thank the member companies of the CAEML IUCRC and the National Science Foundation (award CNS-1624770) for their partial support of this work.